\documentclass{article}

\PassOptionsToPackage{numbers, compress}{natbib}

\usepackage[final]{neurips_2024}

\usepackage[position=bottom]{caption}
\usepackage{amsmath} 
\usepackage[utf8]{inputenc} 
\usepackage[T1]{fontenc}    
\usepackage{url}            
\usepackage{booktabs}       
\usepackage{amsfonts}       
\usepackage{nicefrac}       
\usepackage{microtype}      
\usepackage[numbers]{natbib}
\usepackage[dvipsnames,table,xcdraw]{xcolor}
\definecolor{cvprblue}{rgb}{0.21,0.49,0.74}
\usepackage{graphicx}
\usepackage{caption}
\usepackage{arydshln}
\usepackage{tikz}
\usepackage{wrapfig}
\usepackage{float}
\usepackage{multirow}
\usepackage{pifont}
\usepackage{makecell}
\usepackage{subfigure}
\usepackage{pdfpages}
\usepackage{xcolor}

\usepackage{tikz}
\usepackage{xcolor}

\usepackage{listings}

\usepackage{ulem}

\usepackage[pagebackref,breaklinks,colorlinks,citecolor=cvprblue]{hyperref}

\usepackage{mdframed}
\definecolor{mygray}{gray}{0.97}
\colorlet{shadecolor}{mygray}
\newmdenv[%
  backgroundcolor=mygray, 
  linewidth=0pt
]{newshaded}

\title{Skywork UniPic: Unified Autoregressive Modeling for Visual Understanding and Generation }
\author{
    \textbf{Multimodality Team, Skywork AI} \\
    \texttt{multimodal@skywork.ai}
}

\begin{document}

\maketitle
\vspace{-3em}
\begin{flushleft}
\setlength{\leftskip}{4em}  
    \includegraphics[height=1.5em]{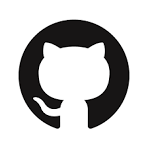}\quad
  {\ttfamily\color{blue}\href{https://github.com/SkyworkAI/UniPic}{https://github.com/SkyworkAI/UniPic}}\\[0.5em]
  \includegraphics[height=1.5em]{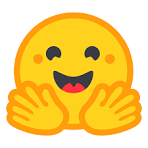}\quad
  {\ttfamily\color{blue}\href{https://huggingface.co/Skywork/Skywork-UniPic-1.5B}{https://huggingface.co/Skywork/Skywork-UniPic-1.5B}}\\[1em]

\end{flushleft}

\begin{abstract}

We introduce Skywork UniPic, a 1.5 billion-parameter autoregressive model that unifies image understanding, text-to-image generation, and image editing within a single architecture—eliminating the need for task-specific adapters or inter-module connectors—and demonstrate that compact multimodal systems can achieve state-of-the-art performance on commodity hardware. Skywork UniPic achieves a GenEval score of 0.86, surpassing most existing unified models; sets a new DPG-Bench complex-generation record of 85.5; attains 5.83 on GEditBench-EN and 3.49 on ImgEdit-Bench for image editing; and generates 1024 × 1024 images with under 15 GB of GPU memory (e.g., RTX 4090). (1) a decoupled encoding strategy that leverages a masked autoregressive encoder for synthesis and a SigLIP2 encoder for understanding, all feeding a shared autoregressive decoder; (2) a progressive, resolution-aware training schedule scaling from 256 × 256 to 1024 × 1024 while dynamically unfreezing parameters to balance capacity and stability; and (3) meticulously curated, 100 million-scale datasets augmented with task-specific reward models to refine generation and editing objectives. By demonstrating that high-fidelity multimodal integration need not incur prohibitive resource demands, Skywork UniPic establishes a practical paradigm for deployable, high-fidelity multimodal AI. Code and weights are publicly available at \url{https://huggingface.co/Skywork/Skywork-UniPic-1.5B}.
\end{abstract}

\section{Introduction}
The rapid evolution of multimodal artificial intelligence has ushered in a paradigm shift toward unified models capable of seamlessly integrating visual perception, generation, and manipulation within a single architectural framework. Recent demonstrations like GPT-4o's\cite{openai2024gpt4o} viral ``Ghiblification'' capability---transforming ordinary photographs into Studio Ghibli-style artworks through natural language interaction---highlight the transformative potential of such systems. These applications reveal a critical limitation in some conventional approaches\cite{pan2025transfermodalitiesmetaqueries,sun2024generativemultimodalmodelsincontext,wang2024illumeilluminatingllmssee}: fragmented pipelines where separate models handle understanding, generation, and editing. Such isolation impedes cross-modal synergy, inflates deployment costs through redundant model stacks, and disrupts natural multi-turn creative workflows. Consequently, the development of natively unified architectures that intrinsically support end-to-end visual comprehension, text-to-image synthesis, and instruction-driven editing has emerged as a pivotal challenge in multimodal artificial intelligence.

\begin{figure}[htbp]
    \centering
    \makebox[\textwidth]{%
        \includegraphics[width=1.\textwidth]{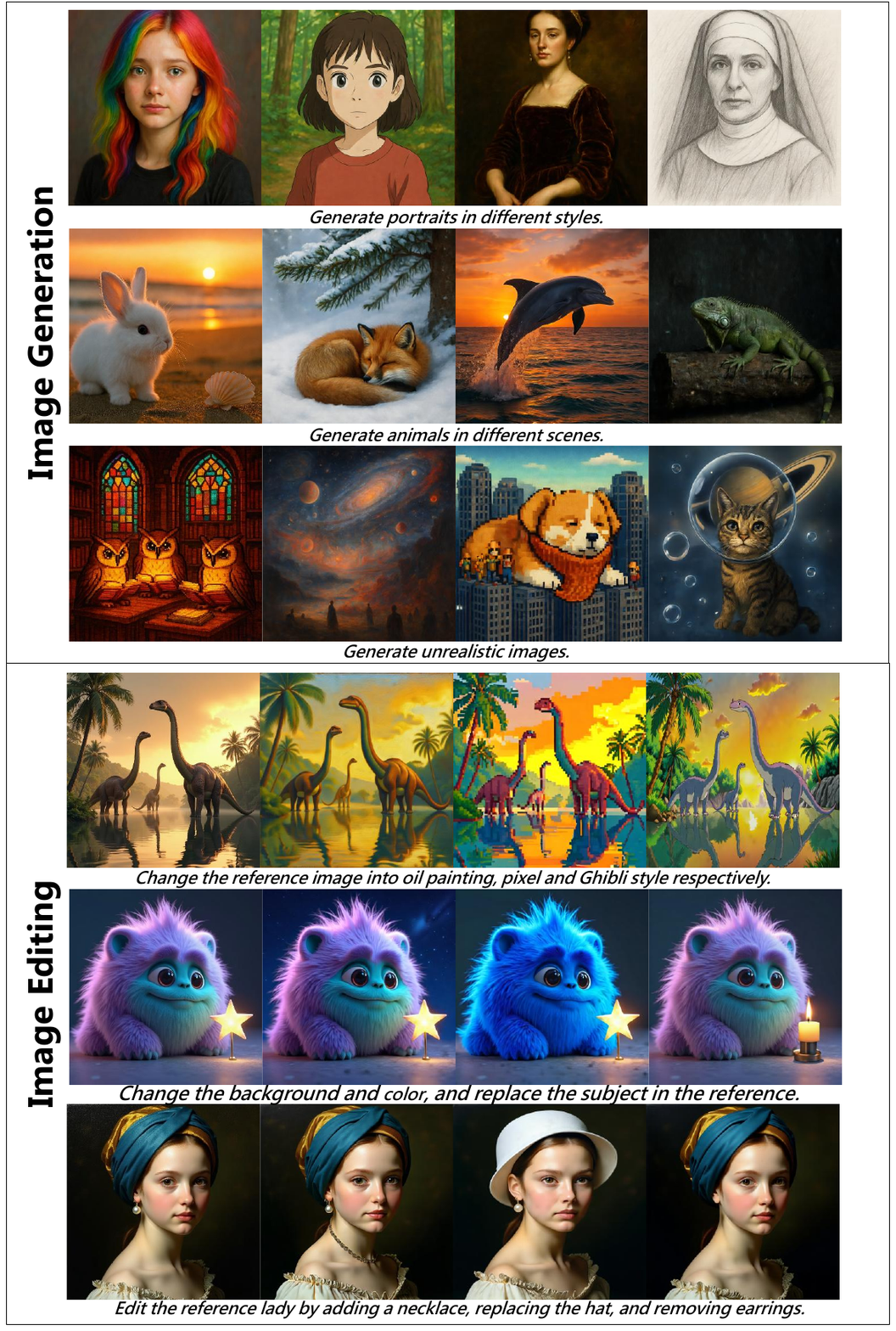}%
    }
    \caption{Showcases of our model's performance on editing and generation tasks.}
    \label{fig:teaser}
\end{figure}

Existing solutions face fundamental constraints. Methods using VQGAN/VAE representations\cite{oord2018neuraldiscreterepresentationlearning,wu2025vilauunifiedfoundationmodel,xie2024showosingletransformerunify,zhou2024transfusionpredicttokendiffuse} prioritize pixel-level reconstruction at the expense of semantic richness, inherently weakening visual understanding capabilities. Alternative approaches\cite{pan2025transfermodalitiesmetaqueries,sun2024generativemultimodalmodelsincontext,wang2024illumeilluminatingllmssee} concatenate pre-trained vision-language and text-to-image models through ad-hoc connectors followed by joint fine-tuning. This piecemeal design fails to achieve deep integration, resulting in performance trade-offs between generation fidelity, editing precision, and reasoning depth. Moreover, prevailing efforts often resort to extreme scaling---deploying multi-billion-parameter models trained on trillion-scale datasets---raising serious concerns about computational efficiency and practical deployability. A crucial question thus remains unanswered: \noindent\textbf{Can a single, parameter-efficient architecture excel simultaneously at visual understanding, high-fidelity image generation, and precise editing, while remaining efficient enough for deployment on commodity hardware?}

We address this challenge through Skywork UniPic, a unified autoregressive model that redefines the efficiency frontier for multimodal integration. The model is built upon a single large language model (LLM), primarily consisting of a MAR encoder, a SigLIP2 encoder, a LLM backbone, and a MAR decoder.  Our architecture fundamentally departs from quantization-based or connector-dependent paradigms by embedding image understanding, text-to-image generation, and image editing within a single end-to-end trainable framework. The core innovation lies in a decoupled visual encoding strategy: we employ the Masked Autoregressive decoder (MAR\cite{li2024autoregressiveimagegenerationvector}) as the backbone for generation-focused representation, optimized for high-fidelity synthesis, while integrating SigLIP2\cite{tschannen2025siglip2multilingualvisionlanguage} for understanding-focused tasks. Critically, both encoders operate within a shared autoregressive objective, enabling bidirectional knowledge transfer where generation enhances visual detail modeling for understanding, and semantic understanding guides coherent editing. This design preserves architectural simplicity while resolving the longstanding tension between pixel-level fidelity and semantic comprehension.

Skywork UniPic achieves unprecedented parameter efficiency without sacrificing capability. With a compact 1.5B language backbone, it establishes new state-of-the-art results across critical benchmarks: surpassing contemporary models on GenEval\cite{ghosh2023genevalobjectfocusedframeworkevaluating} (0.86) for instruction following, achieving 85.5 on DPG-Bench\cite{hu2024ellaequipdiffusionmodels} for complex generation, and leading among unified models on editing tasks (5.83 on GEditBench-EN\cite{liu2025step1xeditpracticalframeworkgeneral}, 3.49 on ImgEdit-Bench\cite{ye2025imgeditunifiedimageediting}), the visualization results as show in Figure\ref{fig:teaser}. Remarkably, it accomplishes this with approximately one-tenth the parameters of comparable systems like BAGEL\cite{deng2025emergingpropertiesunifiedmultimodal} (14B) or UniWorld-V1\cite{lin2025uniworldv1highresolutionsemanticencoders} (19B), while generating 1024$\times$1024 images on consumer-grade hardware (RTX 4090). This efficiency stems from three synergistic innovations: meticulous curation of a hundred-million-scale high-quality dataset emphasizing task balance and semantic diversity; novel text-to-image reward model trained via Group Relative Policy Optimization (GRPO\cite{shao2024deepseekmathpushinglimitsmathematical}) and editing reward model to align with human preferences; and a progressive training curriculum that incrementally introduces task complexity while scaling resolution from 256$^2$ to 1024$^2$.

Our work makes three key contributions to unified multimodal modeling. First, we introduce the natively unified autoregressive architecture that intrinsically supports joint visual understanding, generation, and editing without requiring separate models or connectors, maintaining accessibility for real-world applications. Second, we resolve the semantic-fidelity dichotomy through a decoupled visual encoding strategy that optimizes representation pathways for distinct task requirements while maintaining cross-task synergy. Third, we demonstrate that rigorous data curation, targeted reward modeling, and progressive training enable state-of-the-art performance at unprecedented scale efficiency---proving that high-quality multimodal integration need not demand excessive computational resources. Through extensive validation across four well-known image-related benchmarks and comprehensive ablation studies, we establish Skywork UniPic as a practical foundation for deployable multimodal systems. By open-sourcing the model weights, training code, and technical documentation, we aim to accelerate the adoption of efficient unified vision-language models in resource-constrained environments, bridging the gap between theoretical capability and real-world applicability.

\section{Related Work}

\subsection{Semantic Encoders}
Vision-language models (VLMs) have emerged as the cornerstone of multimodal understanding by introducing semantic encoders that effectively inject visual signals into language models, thereby endowing them with robust image comprehension capabilities. Among these, CLIP\cite{radford2021learningtransferablevisualmodels} established a foundational paradigm through its contrastive learning framework that aligns image and text embeddings in a shared space, enabling remarkable zero-shot classification and retrieval performance. Building on this foundation, SigLIP\cite{zhai2023sigmoidlosslanguageimage} refined the training methodology with a sigmoid-based loss function that eliminated temperature parameter dependencies, enabling more stable scaling. SigLIP2\cite{tschannen2025siglip2multilingualvisionlanguage} integrates multiple advanced techniques—including captioning-based pretraining, self-supervised losses, and online data curation—to produce even richer semantic representations while preserving input aspect ratios across multiple resolutions. These progressive advancements in visual semantic encoding have significantly enhanced zero-shot classification, image-text retrieval, and transfer learning capabilities, establishing crucial foundations for unified models that must balance deep semantic understanding with high-fidelity generation—a balance that remains challenging for existing approaches due to the inherent tension between pixel-level detail preservation and conceptual representation. 
\subsection{Image Generation}

Image generation methods have undergone several distinct architectural paradigms. Early work on Generative Adversarial Networks (GANs\cite{goodfellow2014generativeadversarialnetworks}) demonstrated that adversarial training can produce realistic samples, but often suffered from instability. Diffusion models\cite{ho2020denoisingdiffusionprobabilisticmodels,rombach2022highresolutionimagesynthesislatent,song2021scorebasedgenerativemodelingstochastic}  subsequently introduced a likelihood-based framework, with work such as GLIDE\cite{nichol2022glidephotorealisticimagegeneration}, DALL·E 3\cite{openaidalle3} and Stable Diffusion \cite{rombach2022highresolutionimagesynthesislatent}
achieving high-fidelity, diverse synthesis. More recent diffusion-based variants: LUMINA-Next\cite{zhuo2024luminanextmakingluminat2xstronger},SDXL\cite{podell2023sdxlimprovinglatentdiffusion}, PlayGround v2.5\cite{li2024playgroundv25insightsenhancing},Hunyuan-Dit\cite{li2024hunyuanditpowerfulmultiresolutiondiffusion} and FLUX.1-dev have further optimized image quality and efficiency at scale. In parallel, autoregressive models\cite{llamagen} treat images as sequences of discrete tokens, trading off generation speed for flexibility in conditional synthesis. Latent Diffusion Models (LDMs)\cite{rombach2022highresolutionimagesynthesislatent} have emerged as a practical standard by performing diffusion in a lower-dimensional latent space, thus reducing computation without sacrificing detail. Vector-quantized approaches such as VQGAN\cite{esser2021tamingtransformershighresolutionimage} combine discrete codebooks with adversarial losses to improve perceptual fidelity, although quantization can introduce semantic loss. In contrast, masked autoregressive encoder-decoder (MAR)\cite{li2024autoregressiveimagegenerationvector} operate directly in pixel space using autoregressive masked prediction, eliminating the need for learned codebooks and offering a unified, end-to-end framework that aligns naturally with our autoregressive decoder.

\subsection{Image Editing}

Image editing research has rapidly advanced under natural language supervision, enabling precise and semantically meaningful modifications driven by user instructions. Instruct-Pix2Pix \cite{brooks2023instructpix2pixlearningfollowimage} fine-tunes diffusion models to directly follow edit instructions without additional architectural changes, achieving strong instruction adherence. A pivotal contribution in this direction is Step1X-Edit \cite{liu2025step1xeditpracticalframeworkgeneral}, which established a scalable data generation pipeline across diverse editing tasks and introduced GEdit-Bench for standardized evaluation. Building on this progress, IC-Edit \cite{zhang2025incontexteditenablinginstructional} introduced a context-aware generation mechanism leveraging diffusion Transformers, enabling zero-shot instruction following without architectural modifications, thereby demonstrating strong generalization across unseen editing commands. Concurrently, UltraEdit \cite{zhao2024ultraeditinstructionbasedfinegrainedimage} addressed data scarcity and diversity limitations by constructing a large-scale, automatically curated dataset, significantly improving the quality and fine-grained controllability of language-driven edits. Despite these notable advances, a critical limitation persists: most current systems operate in isolation from broader vision-language understanding and generative modeling pipelines. They typically rely on specialized, standalone architectures that are decoupled from models responsible for image description, reasoning, or synthesis. This architectural fragmentation impedes the realization of seamless, multi-turn interactive workflows, in which users naturally alternate between describing scenes, issuing edit commands, and iteratively refining visual outputs through continuous natural language dialogue. 

\subsection{Unified Models}

Unified multimodal models seek to combine visual understanding and generation within a single architecture, enabling seamless interaction between vision and language. These models can be grouped into four main paradigms: harmonization, decoupling, hybrid, and connector approaches.

The harmonization approach, exemplified by Harmon\cite{wu2025harmonizingvisualrepresentationsunified}, uses a shared MAR\cite{li2024autoregressiveimagegenerationvector} encoder-decoder for both tasks. It builds on findings that MAR representations achieve strong performance in linear probing and respond precisely to visual concepts, suggesting their potential for understanding beyond generation. In contrast, the decoupling strategy, as seen in Janus\cite{wu2024janusdecouplingvisualencoding} and Janus-Pro\cite{chen2025janusprounifiedmultimodalunderstanding}, separates visual encoding into distinct pathways. This design addresses conflicting granularity demands while maintaining a unified Transformer backbone, improving flexibility and task specialization.

Hybrid models like Show-o\cite{xie2024showosingletransformerunify} integrate autoregressive and discrete diffusion mechanisms. This allows support for diverse tasks such as visual question answering, text-to-image generation, and mixed-modal synthesis. Connector-based methods, such as MetaQueries\cite{pan2025transfermodalitiesmetaqueries}, use learnable queries to bridge autoregressive LLMs and diffusion models, enabling modular integration without architectural changes.

Recent advances include BAGEL\cite{deng2025emergingpropertiesunifiedmultimodal}, a large-scale decoder-only model trained on trillions of multimodal tokens. It demonstrates emergent capabilities in multimodal reasoning, including image manipulation, future frame prediction, and 3D navigation. OmniGen2\cite{wu2025omnigen2explorationadvancedmultimodal} introduces separate decoding paths for text and images, along with a decoupled image tokenizer. This design preserves text generation quality while supporting in-context editing and achieving state-of-the-art performance on the OmniContext benchmark.

UniFluid\cite{fan2025unified} adopts a unified autoregressive framework with continuous visual tokens, showing that generation and understanding can mutually benefit under balanced training. Other notable models include BLIP3-o\cite{chen2025blip3ofamilyfullyopen}, which generates CLIP-space features via diffusion Transformers, and OpenUni\cite{wu2025openuni}, a lightweight open-source baseline. Despite significant progress, developing a compact unified model that achieves state-of-the-art performance across understanding, generation, and editing tasks while remaining practical for real-world deployment remains a critical challenge.

\section{Method}
We introduce Skywork UniPic, a unified autoregressive model that natively integrates image understanding, text-to-image generation, and image editing within a single framework. The rapid advancement of multimodal AI has revealed limitations in fragmented approaches where separate specialized models handle different tasks through loosely coupled connectors~\cite{pan2025transfermodalitiesmetaqueries,sun2024generativemultimodalmodelsincontext,wang2024illumeilluminatingllmssee}. Such architectures suffer from suboptimal cross-modal synergy and increased deployment complexity.

Inspired by recent work on unified multimodal modeling, particularly Harmon~\cite{wu2025harmonizingvisualrepresentationsunified} which demonstrated the potential of shared visual representations in autoregressive frameworks, we develop a more sophisticated approach to task unification. While Harmon showed promising results using a single MAR encoder for both understanding and generation, we identify and address a critical limitation: shared encoders can suffer from task interference due to conflicting optimization objectives. Our key insight is that different visual tasks require representations at different levels of granularity—understanding demands semantic richness while generation requires pixel-level fidelity—yet both can benefit from unified processing through a shared language backbone.

Building on this observation, we propose a decoupled encoding strategy within a unified autoregressive framework. Rather than forcing a single encoder to optimize for conflicting objectives, we employ task-specific encoders that feed into a shared language model, enabling both specialized representation learning and cross-task knowledge transfer. This design preserves the benefits of unified training while allowing each encoder to excel at its designated task.

\begin{figure}[htbp]
    \centering
    \includegraphics[width=\textwidth]{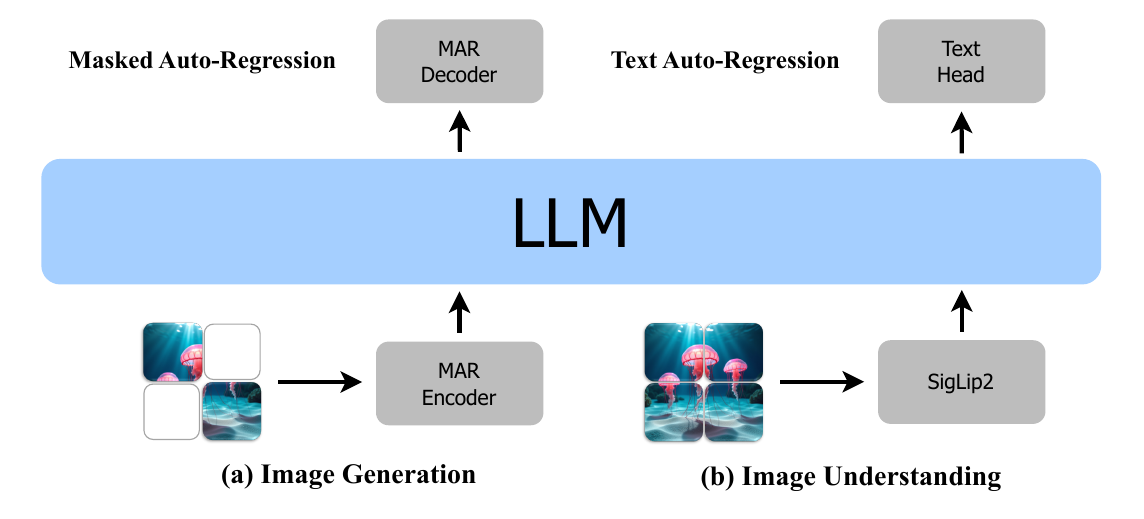}
    \caption{The overall framework of Skywork UniPic. (a) Image generation is achieved through a masked auto-regressive process using the MAR model~\cite{li2024autoregressiveimagegenerationvector}. (b) Image understanding is performed using a SigLIP2 encoder~\cite{tschannen2025siglip2multilingualvisionlanguage} to extract rich visual features, which are subsequently passed to an LLM for autoregressive text generation.  They share a single LLM to promote consistent instruction-following and enable knowledge transfer between generation and understanding tasks}
    \label{fig:model_architecture}
\end{figure}

\subsection{Model Architecture}

Our model consists of four core components: (1) a Masked Autoregressive (MAR) encoder-decoder pair~\cite{li2024autoregressiveimagegenerationvector} for generation-focused visual representation, (2) a SigLIP2 encoder~\cite{tschannen2025siglip2multilingualvisionlanguage} for understanding-focused visual encoding, (3) a shared Qwen2.5-1.5B-Instruct~\cite{qwen2025qwen25technicalreport} language model backbone, and (4) dedicated MLP projection layers that bridge visual encoders to the language model's embedding space, as illustrated in Figure~\ref{fig:model_architecture}.

The architecture departs from the original Harmon framework's shared encoder approach, which we found prone to task interference. Instead, we employ a decoupled encoding strategy where each visual encoder is optimized for its specific task requirements while maintaining unified processing through the shared language backbone. This design preserves the benefits of unified training while allowing task-specific representation learning.

For image generation, we utilize MAR-Huge\footnote{\url{https://huggingface.co/jadechoghari/mar/blob/main/mar-huge.safetensors}} as both encoder and decoder, containing approximately 1B parameters with 20 layers each for encoding and decoding, a hidden dimension of 1280, and 16 attention heads. Images are first encoded into latent representations using a frozen VAE~\cite{rombach2022highresolutionimagesynthesislatent} from the original Harmon framework, preserving low-level visual features and ensuring stable convergence during multimodal training. We scale the generation resolution from 256$\times$256 to 512$\times$512 to enable higher-fidelity synthesis and capture fine-grained visual details, broadening MAR's applicability to high-resolution image synthesis tasks.

For image understanding, we adopt SigLIP2-so400m-patch16-512\footnote{\url{https://huggingface.co/google/siglip2-so400m-patch16-512}} as the visual encoder, leveraging its superior cross-modal alignment capabilities and efficient representation learning demonstrated across vision-language benchmarks. The encoder processes images at 512$\times$512 resolution and extracts semantically rich features optimized for understanding tasks. To further enhance visual understanding capabilities, we continue training based on the SigLIP2-so400m-patch16-512 checkpoint, which provides a solid foundation for cross-modal representation learning.

Two separate two-layer MLPs project the visual encoder outputs to align with the 1.5B language model's embedding space. This separation allows independent optimization of projection mappings for each task while maintaining architectural simplicity and facilitating effective integration with the shared LLM.

\subsection{Training Methodology}

Our training employs a multi-task objective that combines generation and understanding losses:

\begin{itemize}
    \item \textbf{Image Generation} (Diffusion loss):
    \[
    \mathcal{L}_{\text{Gen}} = \mathbb{E}_{\varepsilon, t} \left[ \left\| \varepsilon - \varepsilon_{\theta} \left(x_t \mid t, z \right) \right\|^2 \right]
    \]
    
    \item \textbf{Image Understanding} (Cross-entropy loss):
    \[
    \mathcal{L}_{\text{Und}} = -\frac{1}{N} \sum_{n=1}^{N} \sum_{i=1}^{C} y_{n,i} \log(\hat{y}_{n,i})
    \]
\end{itemize}

These are integrated into the multi-task objective during joint training:
\[
\mathcal{L}_{\text{Total}} = \lambda_{\text{Gen}}\mathcal{L}_{\text{Gen}} + \lambda_{\text{Und}}\mathcal{L}_{\text{Und}}
\]
where $\lambda$ coefficients evolve through training stages to balance task learning dynamics.

We implement a four-stage progressive training curriculum spanning hundred-million-scale pretraining and million-scale supervised fine-tuning. The pipeline begins with \textbf{Stage 1: MAR Pretraining (PT)}, establishing foundational generation capabilities through dedicated training of the MAR encoder-decoder module with particular emphasis on face reconstruction and complex object synthesis. This is followed by \textbf{Stage 2: MAR-LLM Alignment}, where MAR outputs are projected to the LLM embedding space while maintaining frozen LLM parameters, utilizing cosine annealing scheduling to accelerate convergence of the projection layers.

Subsequently, \textbf{Stage 3: Joint Optimization (CT)} unfreezes the LLM for cross-modal tuning under the multi-task objective $\mathcal{L}_{\text{Total}}$ with loss weights $\lambda_{\text{Gen}}=1$ and $\lambda_{\text{Und}}=0.01$, yielding 12-15\% improvements in instruction adherence metrics. The process concludes with \textbf{Stage 4: Supervised Fine-tuning (SFT)}, which refines the unified model using reward-filtered samples with quality threshold above 0.9, incorporating the full $\mathcal{L}_{\text{Total}}$ objective with editing loss components to polish final task performance.

Resolution scaling occurs progressively from $256^2$ in early stages to $1024^2$ in final training, with generation tasks reaching 1024$\times$1024 and understanding tasks stabilizing at 512$\times$512. This staged approach allows the model to learn fundamental capabilities at lower resolutions before adapting to high-resolution synthesis requirements.


\begin{table*}[ht]
\centering
\caption{UniPic Training Configuration Across Learning Stages}
\begin{tabular}{lcccc}
\toprule
\textbf{Hyperparameter} & \textbf{PT} & \textbf{Alignment} & \textbf{CT} & \textbf{SFT} \\
\midrule
Learning rate & $5.0 \times 10^{-5}$ & $1 \times 10^{-5}$ & $1.0 \times 10^{-5}$ & $5 \times 10^{-6}$ \\
LR scheduler & Constant & Cosine decay & Cosine decay & Cosine decay \\
Weight decay & 0.0 & 0.02 & 0.02 & 0.02 \\
Gradient clipping & 1.0 & 1.0 & 1.0 & 1.0 \\
Optimizer & \multicolumn{4}{c}{AdamW ($\beta_1=0.9$, $\beta_2=0.95$, $\epsilon=10^{-15}$)} \\
Loss weights (U:G:E)\footnotemark & 0:1:0 & -- & 0.01:1:1 & 0.01:1:1 \\
Warmup ratio & 0.05 & 0.05 & 0.01 & 0.01 \\
Training epochs & 800 & 3 & 3 & 2 \\
EMA decay & 0.9999 & -- & 0.9999 & 0.995 \\
Training samples & 130M & 130M & 130M & 3M \\
\multicolumn{5}{l}{\textit{Image resolution (width $\times$ height)}} \\
\quad Generation & $512\times512$ & $1024\times1024$ & $1024\times1024$ & $1024\times1024$ \\
\quad Understanding & $256\times256$ & $512\times512$ & $512\times512$ & $512\times512$ \\
\bottomrule
\end{tabular}
\footnotetext{U:G:E = Understanding:Generation:Editing loss weights}
\label{tab:training-recipe-full}
\end{table*}
\footnotetext{U:G:E = Understanding:Generation:Editing loss weights}

Hyperparameters are detailed in Table~\ref{tab:training-recipe-full}. The training stack utilizes \texttt{bf16} mixed-precision and is optimized with DeepSpeed ZeRO-3~\cite{aminabadi2022deepspeedinferenceenablingefficient}. We use a global batch size of 4096 for pre-training (PT) and 512 for supervised fine-tuning (SFT). The model architecture consists of an 800M parameter MAR module combined with a 1.5B parameter language model backbone.

\subsection{Data Quality Assurance}
\label{sec:data_quality}

To ensure training data quality, we develop two specialized reward models based on Qwen-VL architecture~\cite{bai2023qwenvlversatilevisionlanguagemodel}: \textbf{Skywork-ImgReward} for visual quality assessment and \textbf{Skywork-EditReward} for image editing accuracy evaluation.

\textbf{Skywork-ImgReward} is trained using Group Relative Policy Optimization (GRPO)~\cite{shao2024deepseekmathpushinglimitsmathematical}, leveraging a custom-designed \textit{paired ranking reward function} that combines learned pairwise ranking scores ($r_\theta$) with format-based scores ($r_{\mathrm{format}}$):
\begin{equation}
r(x, y_i) = \underbrace{r_\theta(x, y_i)}_{\text{pairwise ranking}} + \underbrace{r_{\mathrm{format}}(x, y_i)}_{\text{format reward}}
\end{equation}

Training data integrates several public datasets, including Pick-a-Pic~\cite{kirstain2023pickapicopendatasetuser}, ImageRewardDB~\cite{xu2023imagereward}, and HPSv2~\cite{wu2023humanpreferencescorev2}, augmented with curated samples focused on human figure quality.

\textbf{Skywork-EditReward} is trained via supervised fine-tuning on high-quality editing datasets including HumanEdit~\cite{bai2025humanedithighqualityhumanrewardeddataset}, UltraEdit~\cite{zhao2024ultraeditinstructionbasedfinegrainedimage}, and SuperEdit-40K~\cite{li2025supereditrectifyingfacilitatingsupervision}, enabling fine-grained assessment of instruction alignment and semantic correctness in image edits.

Our data curation pipeline applies rigorous filtration, first discarding samples with reward scores below 0.9, then employing multi-check mechanisms using VQAScore~\cite{lin2024evaluatingtexttovisualgenerationimagetotext} as additional quality heuristic. Analysis reveals four primary failure modes: instruction-alignment deviations, visual artifacts, semantic inconsistencies, and edit non-compliance. This curated dataset ensures high data homogeneity and enhances model generalization across diverse visual categories including human figures, animals, and text rendering.

\section{Main Results}

\subsection{Evaluation Setup}

To comprehensively assess the unified capabilities of Skywork UniPic, we adopt a multi-faceted evaluation strategy encompassing image understanding, text-to-image generation, and image editing across established benchmarks.

\paragraph{Benchmarks.} For text-to-image generation, we evaluate on GenEval\cite{ghosh2023genevalobjectfocusedframeworkevaluating} which measures compositional understanding and object-focused alignment, and DPG-Bench\cite{hu2024ellaequipdiffusionmodels} which assesses complex instruction following and long prompt adherence capabilities. These benchmarks capture both fine-grained compositional reasoning and general-purpose generation quality.

Image editing capabilities are assessed using GEdit-Bench-EN\cite{liu2025step1xeditpracticalframeworkgeneral} and ImgEdit-Bench\cite{ye2025imgeditunifiedimageediting} as primary evaluation suites. Built from authentic user requests covering diverse editing scenarios, these benchmarks closely mirror practical editing needs and provide comprehensive coverage of instruction-based image modification tasks including object addition/removal, style transfer, and attribute modification.

\paragraph{Evaluation Protocol}
All image generation tasks employ 64 sampling steps with 1024$\times$1024 resolution outputs and classifier-free guidance scale of 3 for optimal quality-diversity trade-off. Performance assessment utilizes official benchmark scripts and automated evaluation metrics, with all scores reported from single evaluation runs without reranking or multi-sampling to ensure reproducible results.

\paragraph{Baselines}

We compare against several categories of state-of-the-art models. Unified models include OmniGen/OmniGen2~\cite{xiao2024omnigenunifiedimagegeneration, wu2025omnigen2explorationadvancedmultimodal}, Janus/Janus-Pro~\cite{wu2024janusdecouplingvisualencoding, chen2025janusprounifiedmultimodalunderstanding}, BAGEL~\cite{deng2025emergingpropertiesunifiedmultimodal}, UniWorld-V1~\cite{lin2025uniworldv1highresolutionsemanticencoders}, Show-o~\cite{xie2024showosingletransformerunify}, BLIP3-o~\cite{chen2025blip3ofamilyfullyopen}, MetaQuery-XL~\cite{pan2025transfermodalitiesmetaqueries}, and Ovis-U1~\cite{wang2025ovisu1technicalreport}. 

Specialized generation models comprise diffusion approaches (FLUX.1-dev~\cite{yang2024158bitflux}, SD3-medium~\cite{stabilityai2025sd3medium}, SDXL~\cite{podell2023sdxlimprovinglatentdiffusion}, DALL-E 3~\cite{openaidalle3}, LUMINA-Next~\cite{zhuo2024luminanextmakingluminat2xstronger}, Hunyuan-DiT~\cite{li2024hunyuanditpowerfulmultiresolutiondiffusion}, PixArt-$\sum$~\cite{chen2023pixartalphafasttrainingdiffusion}, NOVA~\cite{jiang2025novagenerativelanguagemodels}) and autoregressive models (TokenFlow-XL~\cite{qu2024tokenflowunifiedimagetokenizer}, Emu3-Gen~\cite{wang2024emu3nexttokenpredictionneed}). For editing, we compare against Step1X-Edit~\cite{liu2025step1xeditpracticalframeworkgeneral}, ICEdit~\cite{zhang2025incontexteditenablinginstructional}, AnyEdit~\cite{jiang2025anyediteditknowledgeencoded}, UltraEdit~\cite{zhao2024ultraeditinstructionbasedfinegrainedimage}, Instruct-Pix2Pix~\cite{brooks2023instructpix2pixlearningfollowimage}, and MagicBrush~\cite{zhang2024magicbrushmanuallyannotateddataset}. Proprietary models include GPT-4o~\cite{openai2025gpt4o} and Gemini-2.0-flash~\cite{gemini2flash}.

\begin{figure}[htbp]
    \centering
    \includegraphics[width=\textwidth]{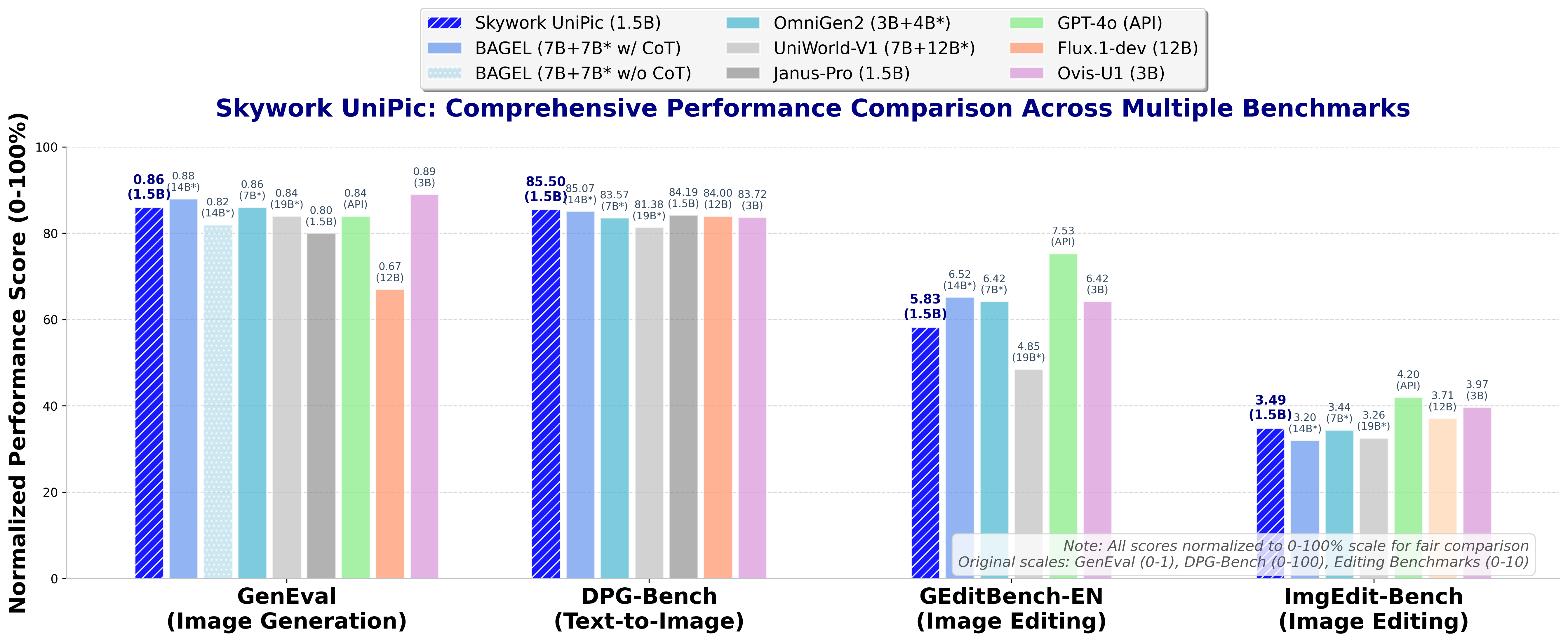}
    \caption{Performance comparison across multiple benchmarks. Skywork UniPic demonstrates competitive performance across understanding, generation, editing, and in-context tasks while maintaining exceptional parameter efficiency with only 1.5B activated parameters.}
    \label{fig:performance_comparison}
\end{figure}

Despite utilizing only 1.5B activated parameters, Skywork UniPic demonstrates competitive or superior performance compared to significantly larger unified models (typically 7B+ parameters), highlighting the effectiveness of our architectural design and training methodology.The corresponding performance metrics for each task are summarized in Figure \ref{fig:performance_comparison}.

\subsection{Text-to-Image Generation}

We assess Skywork UniPic's T2I generation capabilities on two standard benchmarks: GenEval and DPG-Bench, which evaluate compositional understanding and long prompt following respectively. Our model demonstrates highly competitive performance, particularly when considering its resource efficiency.

\textbf{Evaluation on GenEval.} As shown in Table~\ref{tab:geneval_results}, Skywork UniPic achieves an overall score of 0.86 on GenEval, demonstrating strong compositional understanding across diverse generation tasks. The model performs particularly well on single object generation (98.44\%) and two object composition (92.42\%), while maintaining solid performance on color understanding (90.69\%) and spatial positioning (89.00\%). Counting tasks (74.06\%) and color attribution (72.25\%) present greater challenges, consistent with observations across unified models in the literature.

\begin{table}[htbp]
\centering
\caption{Comprehensive comparison on GenEval benchmark. † denotes using rewritten prompts.}
\label{tab:geneval_results}
\begin{tabular}{lcccccccc}
\toprule
\textbf{Model} & \textbf{Single} & \textbf{Two} & \textbf{Count} & \textbf{Color} & \textbf{Position} & \textbf{Attr} & \textbf{Overall} \\
\midrule
\multicolumn{8}{l}{\textit{Diffusion Models}} \\
SDv2.1\cite{rombach2022highresolutionimagesynthesislatent} & 0.98 & 0.51 & 0.44 & 0.85 & 0.07 & 0.17 & 0.50 \\
SDXL\cite{podell2023sdxlimprovinglatentdiffusion} & 0.98 & 0.74 & 0.39 & 0.85 & 0.15 & 0.23 & 0.55 \\
IF-XL & 0.97 & 0.74 & 0.66 & 0.81 & 0.13 & 0.35 & 0.61 \\
LUMINA-Next\cite{zhuo2024luminanextmakingluminat2xstronger} & 0.92 & 0.46 & 0.48 & 0.70 & 0.09 & 0.13 & 0.46 \\
SD3-medium\cite{stabilityai2025sd3medium} & 0.99 & 0.94 & 0.72 & 0.89 & 0.33 & 0.60 & 0.74 \\
FLUX.1-dev\cite{yang2024158bitflux} & 0.99 & 0.81 & 0.79 & 0.74 & 0.20 & 0.47 & 0.67 \\
NOVA\cite{jiang2025novagenerativelanguagemodels} & 0.99 & 0.91 & 0.62 & 0.85 & 0.33 & 0.56 & 0.71 \\
\midrule
\multicolumn{8}{l}{\textit{Autoregressive Models}} \\
TokenFlow-XL\cite{qu2024tokenflowunifiedimagetokenizer} & 0.95 & 0.60 & 0.41 & 0.81 & 0.16 & 0.24 & 0.55 \\
Janus\cite{wu2024janusdecouplingvisualencoding} & 0.97 & 0.68 & 0.30 & 0.84 & 0.46 & 0.42 & 0.61 \\
Janus Pro\cite{chen2025janusprounifiedmultimodalunderstanding} & 0.99 & 0.89 & 0.59 & 0.90 & 0.79 & 0.66 & 0.80 \\
Emu3-Gen\cite{wang2024emu3nexttokenpredictionneed} & 0.99 & 0.81 & 0.42 & 0.80 & 0.49 & 0.45 & 0.66 \\
Show-o\cite{xie2024showosingletransformerunify} & 0.98 & 0.80 & 0.66 & 0.84 & 0.31 & 0.50 & 0.68 \\
\midrule
\multicolumn{8}{l}{\textit{Unified Models}} \\
OmniGen\cite{xiao2024omnigenunifiedimagegeneration} & 0.98 & 0.84 & 0.66 & 0.74 & 0.40 & 0.43 & 0.68 \\
OmniGen2\cite{wu2025omnigen2explorationadvancedmultimodal} & 1.00 & 0.95 & 0.64 & 0.88 & 0.55 & 0.76 & 0.80 \\
OmniGen2$^\dagger$ & 0.99 & 0.96 & 0.74 & 0.98 & 0.71 & 0.75 & 0.86 \\
MetaQuery-XL$^\dagger$\cite{pan2025transfermodalitiesmetaqueries} & - & - & - & - & - & - & 0.80 \\
BLIP3-o$^\dagger$ 4B\cite{chen2025blip3ofamilyfullyopen} & - & - & - & - & - & - & 0.81 \\
BLIP3-o$^\dagger$ 8B & - & - & - & - & - & - & 0.84 \\
BAGEL\cite{deng2025emergingpropertiesunifiedmultimodal} & 0.99 & 0.94 & 0.81 & 0.88 & 0.64 & 0.63 & 0.82 \\
BAGEL$^\dagger$ & 0.98 & 0.95 & 0.84 & 0.95 & 0.78 & 0.77 & 0.88 \\
UniWorld-V1\cite{lin2025uniworldv1highresolutionsemanticencoders} & 0.99 & 0.93 & 0.79 & 0.89 & 0.49 & 0.70 & 0.80 \\
UniWorld-V1$^\dagger$ & 0.98 & 0.93 & 0.81 & 0.89 & 0.74 & 0.71 & 0.84 \\
Ovis-U1\cite{wang2025ovisu1technicalreport} & 0.98 & 0.98 & 0.90 & 0.92 & 0.79 & 0.75 & 0.89 \\
\midrule
\multicolumn{8}{l}{\textit{Proprietary Models}} \\
GPT-4o\cite{openai2025gpt4o} & 0.99 & 0.92 & 0.85 & 0.92 & 0.75 & 0.61 & 0.84 \\
\midrule
\textbf{Skywork UniPic} & \textbf{0.98} & \textbf{0.92} & \textbf{0.74} & \textbf{0.91} & \textbf{0.89} & \textbf{0.72} & \textbf{0.86} \\
\bottomrule
\end{tabular}
\end{table}

\textbf{Evaluation on DPG-Bench.} On DPG-Bench, Skywork UniPic achieves an overall score of 85.5, demonstrating competitive performance in long prompt following and complex scene understanding. Table~\ref{tab:dpg_results} shows detailed comparisons across different evaluation categories, where our model maintains consistent performance across global coherence, entity recognition, attribute understanding, and relational reasoning.

These results are particularly notable given our model's compact 1.5B parameter count compared to significantly larger unified alternatives like BAGEL (14B) or UniWorld-V1 (19B), highlighting the effectiveness of our decoupled encoding strategy and progressive training methodology.

\begin{table}[htbp]
\centering
\caption{Comprehensive comparison on DPG-Bench across different semantic categories.}
\label{tab:dpg_results}
\begin{tabular}{lcccccc}
\toprule
\textbf{Model} & \textbf{Global} & \textbf{Entity} & \textbf{Attribute} & \textbf{Relation} & \textbf{Other} & \textbf{Overall} \\
\midrule
\multicolumn{7}{l}{\textit{Diffusion Models}} \\
LUMINA-Next\cite{zhuo2024luminanextmakingluminat2xstronger} & 82.82 & 88.65 & 86.44 & 80.53 & 81.82 & 74.63 \\
SDXL\cite{podell2023sdxlimprovinglatentdiffusion} & 83.27 & 82.43 & 80.91 & 86.76 & 80.41 & 74.65 \\
PlayGroundv2.5\cite{li2024playgroundv25insightsenhancing} & 83.06 & 82.59 & 81.20 & 84.08 & 83.50 & 75.47 \\
Hunyuan-DiT\cite{li2024hunyuanditpowerfulmultiresolutiondiffusion} & 84.59 & 80.59 & 88.01 & 74.36 & 86.41 & 78.87 \\
PixArt-$\sum$\cite{chen2023pixartalphafasttrainingdiffusion} & 86.89 & 82.89 & 88.94 & 86.59 & 87.68 & 80.54 \\
DALLE3\cite{openaidalle3} & 90.97 & 89.61 & 88.39 & 90.58 & 89.83 & 83.50 \\
SD3-medium\cite{stabilityai2025sd3medium} & 87.90 & 91.01 & 88.83 & 80.70 & 88.68 & 84.08 \\
FLUX.1-dev\cite{yang2024158bitflux} & 82.10 & 89.50 & 88.70 & 91.10 & 89.40 & 84.00 \\
\midrule
\multicolumn{7}{l}{\textit{Autoregressive Models}} \\
Show-o\cite{xie2024showosingletransformerunify} & 79.33 & 75.44 & 78.02 & 84.45 & 60.80 & 67.27 \\
EMU3\cite{wang2024emu3nexttokenpredictionneed} & 85.21 & 86.68 & 86.84 & 90.22 & 83.15 & 80.60 \\
TokenFlow-XL\cite{qu2024tokenflowunifiedimagetokenizer} & 78.72 & 79.22 & 81.29 & 85.22 & 71.20 & 73.38 \\
Janus\cite{wu2024janusdecouplingvisualencoding} & 82.33 & 87.38 & 87.70 & 85.46 & 86.41 & 79.68 \\
Janus Pro\cite{chen2025janusprounifiedmultimodalunderstanding} & 86.90 & 88.90 & 89.40 & 89.32 & 89.48 & 84.19 \\
BLIP3-o 4B\cite{chen2025blip3ofamilyfullyopen} & - & - & - & - & - & 79.36 \\
BLIP3-o 8B & - & - & - & - & - & 81.60 \\
\midrule
\multicolumn{7}{l}{\textit{Unified Models}} \\
OmniGen\cite{xiao2024omnigenunifiedimagegeneration} & 87.90 & 88.97 & 88.47 & 87.95 & 83.56 & 81.16 \\
OmniGen2\cite{wu2025omnigen2explorationadvancedmultimodal} & 88.81 & 88.83 & 90.18 & 89.37 & 90.27 & 83.57 \\
BAGEL\cite{deng2025emergingpropertiesunifiedmultimodal} & 88.94 & 90.37 & 91.29 & 90.82 & 88.67 & 85.07 \\
UniWorld-V1\cite{lin2025uniworldv1highresolutionsemanticencoders} & 83.64 & 88.39 & 88.44 & 89.27 & 87.22 & 81.38 \\
Ovis-U1\cite{wang2025ovisu1technicalreport} & 82.37 & 90.08 & 88.68 & 93.35 & 85.20 & 83.72 \\
\midrule
\textbf{Skywork UniPic} & \textbf{89.65} & \textbf{87.78} & \textbf{90.84} & \textbf{91.89} & \textbf{91.95} & \textbf{85.50} \\
\bottomrule
\end{tabular}
\end{table}

\subsection{Image Editing}

Image editing represents a core strength of Skywork UniPic's unified architecture. We evaluate the model's editing capabilities on both GEdit-Bench and ImgEdit-Bench, which assess instruction-based image modification across diverse scenarios.

\textbf{Evaluation on GEdit-Bench.} As demonstrated in Table~\ref{tab:gedit_results}, Skywork UniPic achieves strong performance with an overall score of 5.83, placing it among the top-tier unified models. The model demonstrates particular strength in semantic consistency (SC) with a score of 6.72, indicating robust instruction-following capabilities. While perceptual quality (PQ) scores show room for improvement at 6.18, the model's ability to make precise, localized edits while preserving unmodified regions demonstrates the effectiveness of our unified architecture.

\begin{table}[htbp]
\centering
\caption{Comprehensive comparison on GEdit-Bench-EN showing semantic consistency (SC) and perceptual quality (PQ) metrics. Higher scores are better for all metrics.}
\label{tab:gedit_results}
\begin{tabular}{lccc}
\toprule
\textbf{Model} & \textbf{SC $\uparrow$} & \textbf{PQ $\uparrow$} & \textbf{Overall $\uparrow$} \\
\midrule
\multicolumn{4}{l}{\textit{Proprietary Models}} \\
Gemini-2.0-flash\cite{gemini2flash} & 6.73 & 6.61 & 6.32 \\
GPT-4o\cite{openai2025gpt4o} & 7.85 & 7.62 & 7.53 \\
\midrule
\multicolumn{4}{l}{\textit{Specialized Editing Models}} \\
Instruct-Pix2Pix\cite{brooks2023instructpix2pixlearningfollowimage} & 3.58 & 5.49 & 3.68 \\
MagicBrush\cite{zhang2024magicbrushmanuallyannotateddataset} & 4.68 & 5.66 & 4.52 \\
AnyEdit\cite{jiang2025anyediteditknowledgeencoded} & 3.18 & 5.82 & 3.21 \\
ICEdit\cite{zhang2025incontexteditenablinginstructional} & 5.11 & 6.85 & 4.84 \\
Step1X-Edit\cite{liu2025step1xeditpracticalframeworkgeneral} & 7.09 & 6.76 & 6.70 \\
\midrule
\multicolumn{4}{l}{\textit{Unified Models}} \\
OmniGen\cite{xiao2024omnigenunifiedimagegeneration} & 5.96 & 5.89 & 5.06 \\
OmniGen2\cite{wu2025omnigen2explorationadvancedmultimodal} & 7.16 & 6.77 & 6.41 \\
BAGEL\cite{deng2025emergingpropertiesunifiedmultimodal} & 7.36 & 6.83 & 6.52 \\
UniWorld-V1\cite{lin2025uniworldv1highresolutionsemanticencoders} & 4.93 & 7.43 & 4.85 \\
Ovis-U1\cite{wang2025ovisu1technicalreport} & - & - & 6.42\\
\midrule
\textbf{Skywork UniPic} & \textbf{6.72} & \textbf{6.18} & \textbf{5.83} \\
\bottomrule
\end{tabular}
\end{table}

\textbf{Evaluation on ImgEdit-Bench.} To further validate our model's editing capabilities across diverse scenarios, we evaluate Skywork UniPic on ImgEdit-Bench, a comprehensive benchmark covering nine distinct editing categories. As demonstrated in Table~\ref{tab:imgedit_results}, Skywork UniPic achieves competitive performance with an overall score of 3.49, establishing itself among the leading unified models in comprehensive image editing evaluation.

The results reveal noteworthy patterns in our model's performance across different editing categories. Skywork UniPic demonstrates particularly strong capabilities in Action editing (4.04) and Style modification (4.76), benefiting from our progressive training methodology that emphasizes multi-stage capability development and comprehensive data curation across diverse editing scenarios. The model also shows solid performance in Background editing (3.77) and Replace operations (4.31), indicating robust understanding of spatial relationships and object substitution.

Compared to other unified models, Skywork UniPic outperforms OmniGen (2.96) and approaches the performance of leading specialized editing models like ICEdit (3.05) and Step1X-Edit (3.06), while maintaining the advantage of unified architecture that handles multiple modalities within a single framework. The superior performance of BAGEL (3.20) and UniWorld-V1 (3.26) on certain categories demonstrates the benefits of larger parameter scales and extensive training data, yet our model achieves comparable results with significantly fewer parameters, highlighting the efficiency of our architectural design and training strategy.

\begin{table}[htbp]
\centering
\caption{Comprehensive comparison on ImgEdit-Bench showing performance across nine editing categories. Higher scores are better for all metrics.}
\label{tab:imgedit_results}
\scalebox{0.75}{%
\begin{tabular}{lccccccccccc}
\toprule
\textbf{Model} & \textbf{Add} & \textbf{Adjust} & \textbf{Extract} & \textbf{Replace} & \textbf{Remove} & \textbf{Background} & \textbf{Style} & \textbf{Hybrid} & \textbf{Action} & \textbf{Overall} \\
\midrule
\multicolumn{11}{l}{\textit{Proprietary Models}} \\
GPT-4o\cite{openai2025gpt4o} & 4.61 & 4.33 & 2.90 & 4.35 & 3.66 & 4.57 & 4.93 & 3.96 & 4.89 & 4.20 \\
\midrule
\multicolumn{11}{l}{\textit{Specialized Editing Models}} \\
MagicBrush\cite{zhang2024magicbrushmanuallyannotateddataset} & 2.84 & 1.58 & 1.51 & 1.97 & 1.58 & 1.75 & 2.38 & 1.62 & 1.22 & 1.90 \\
Instruct-Pix2Pix\cite{brooks2023instructpix2pixlearningfollowimage} & 2.45 & 1.83 & 1.44 & 2.01 & 1.50 & 1.44 & 3.55 & 1.20 & 1.46 & 1.88 \\
AnyEdit\cite{jiang2025anyediteditknowledgeencoded} & 3.18 & 2.95 & 1.88 & 2.47 & 2.23 & 2.24 & 2.85 & 1.56 & 2.65 & 2.45 \\
UltraEdit\cite{zhao2024ultraeditinstructionbasedfinegrainedimage} & 3.44 & 2.81 & 2.13 & 2.96 & 1.45 & 2.83 & 3.76 & 1.91 & 2.98 & 2.70 \\
Step1X-Edit\cite{liu2025step1xeditpracticalframeworkgeneral} & 3.88 & 3.14 & 1.76 & 3.40 & 2.41 & 3.16 & 4.63 & 2.64 & 2.52 & 3.06 \\
ICEdit\cite{zhang2025incontexteditenablinginstructional} & 3.58 & 3.39 & 1.73 & 3.15 & 2.93 & 3.08 & 3.84 & 2.04 & 3.68 & 3.05 \\
\midrule
\multicolumn{11}{l}{\textit{Unified Models}} \\
OmniGen\cite{xiao2024omnigenunifiedimagegeneration} & 3.47 & 3.04 & 1.71 & 2.94 & 2.43 & 3.21 & 4.19 & 2.24 & 3.38 & 2.96 \\
OmniGen2\cite{wu2025omnigen2explorationadvancedmultimodal} & 3.57 & 3.06 & 1.77 & 3.74 & 3.20 & 3.57 & 4.81 & 2.52 & 4.68 & 3.44 \\
BAGEL\cite{deng2025emergingpropertiesunifiedmultimodal} & 3.56 & 3.31 & 1.70 & 3.30 & 2.62 & 3.24 & 4.49 & 2.38 & 4.17 & 3.20 \\
UniWorld-V1\cite{lin2025uniworldv1highresolutionsemanticencoders} & 3.82 & 3.64 & 2.27 & 3.47 & 3.24 & 2.99 & 4.21 & 2.96 & 2.74 & 3.26 \\
Ovis-U1\cite{wang2025ovisu1technicalreport} & 4.13 & 3.62 & 2.98 & 4.45 & 4.06 & 4.22 & 4.69 & 3.45 & 4.61 & 4.00 \\
\midrule
\textbf{Skywork UniPic} & \textbf{3.66} & \textbf{3.51} & \textbf{2.06} & \textbf{4.31} & \textbf{2.77} & \textbf{3.77} & \textbf{4.76} & \textbf{2.56} & \textbf{4.04} & \textbf{3.49} \\
\bottomrule
\end{tabular}
}
\end{table}

\subsection{Qualitative Results}

\begin{figure}[htbp]
    \centering
    \vspace{-3em}
    \makebox[\textwidth]{\includegraphics[width=1.2\textwidth]{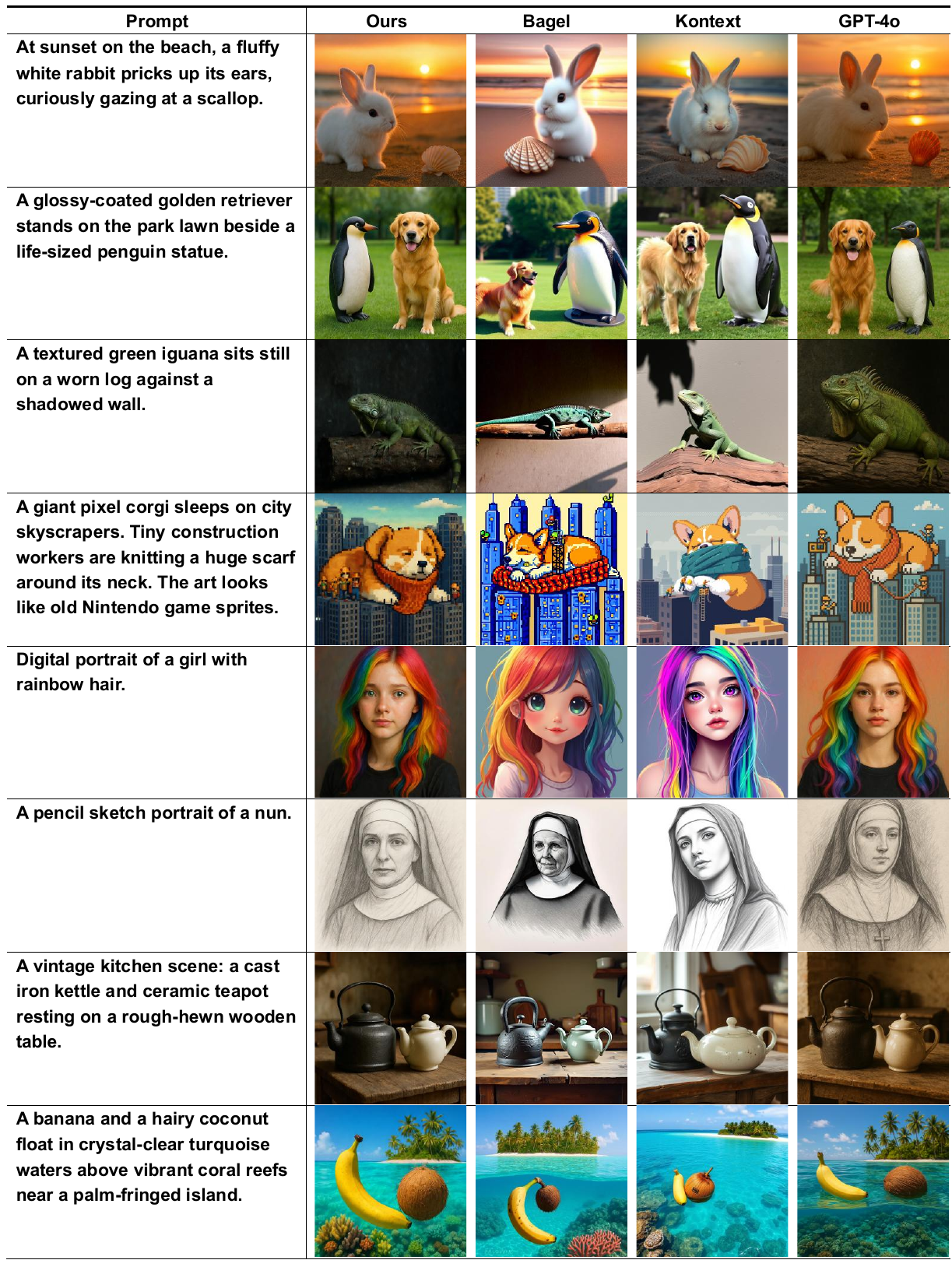}}
    \vspace{-1em}
    \caption{Qualitative comparison of text-to-image generation results. Skywork UniPic produces high-quality images that accurately reflect textual prompts while maintaining competitive visual fidelity compared to both open-source and proprietary models.}
    \label{fig:t2i_qualitative}
\end{figure}

\textbf{Text-to-Image Generation Quality.} Figure~\ref{fig:t2i_qualitative} presents qualitative comparisons between Skywork UniPic and both open-source and proprietary models on text-to-image generation tasks. Our model demonstrates competitive visual quality and strong adherence to textual prompts across diverse scenarios, from simple object generation to complex scene composition. The results show that despite its compact size, Skywork UniPic produces images with comparable fidelity and semantic accuracy to much larger specialized models.

\begin{figure}[htbp]
    \centering
    \vspace{-3em}
    \makebox[\textwidth]{\includegraphics[width=1.2\textwidth]{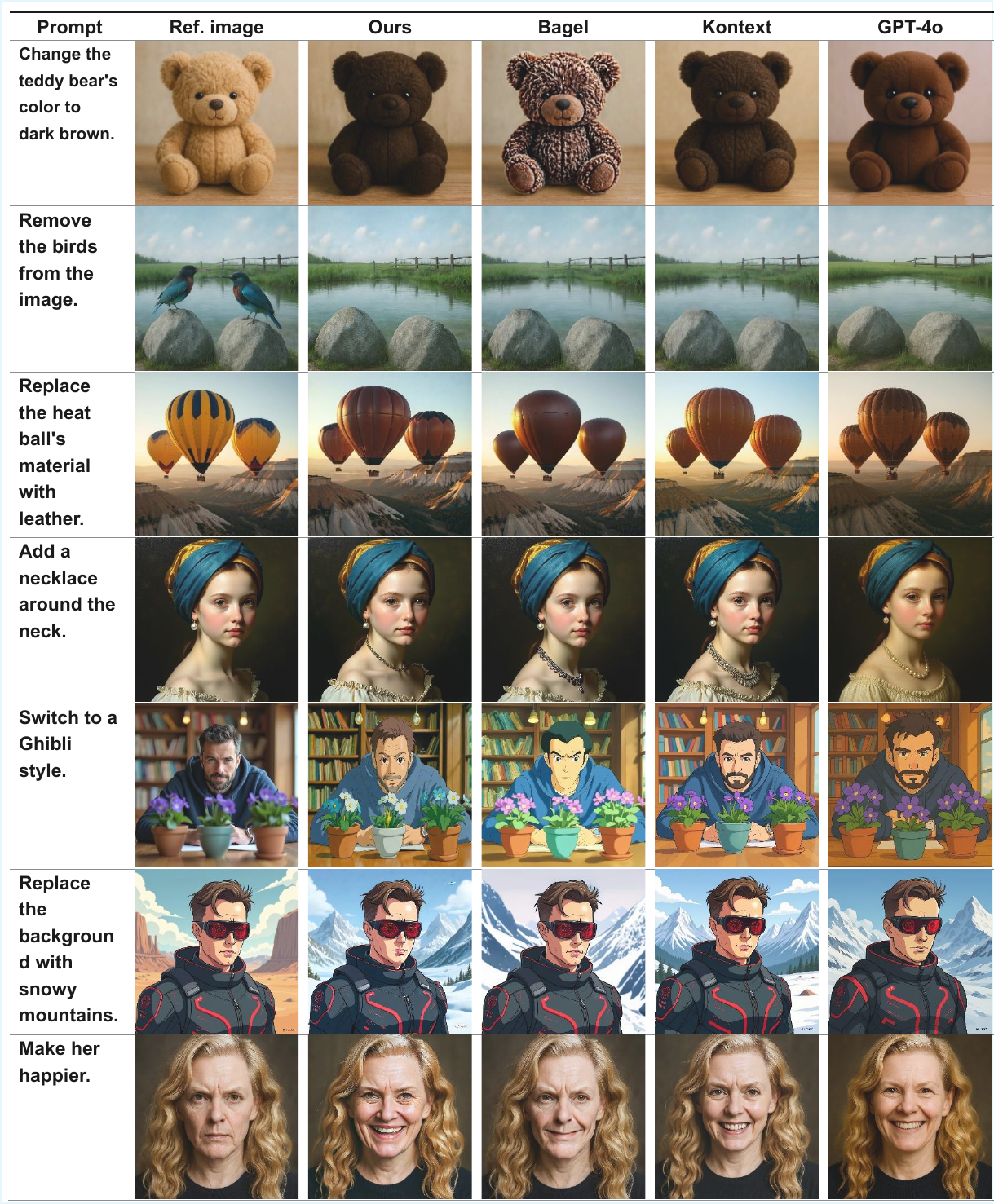}}
    \vspace{-1em}
    \caption{Qualitative comparison of image editing results. Skywork UniPic successfully handles diverse editing instructions while preserving image quality and maintaining consistency in unmodified regions, demonstrating the effectiveness of our unified approach.}
    \label{fig:editing_qualitative}
\end{figure}

\textbf{Image Editing Capabilities.} Figure~\ref{fig:editing_qualitative} showcases Skywork UniPic's image editing performance compared to state-of-the-art editing models. The model demonstrates precise instruction following across various editing scenarios, including object addition/removal, style transfer, attribute modification, and complex compositional changes. Notably, the model maintains consistency in unedited regions while accurately implementing the requested modifications, highlighting the benefits of our unified architecture approach.

\section{Limitation and Discussion}

\begin{figure}[htbp]
    \centering
    \vspace{-3em}
    \makebox[\textwidth]{\includegraphics[width=1.2\textwidth]{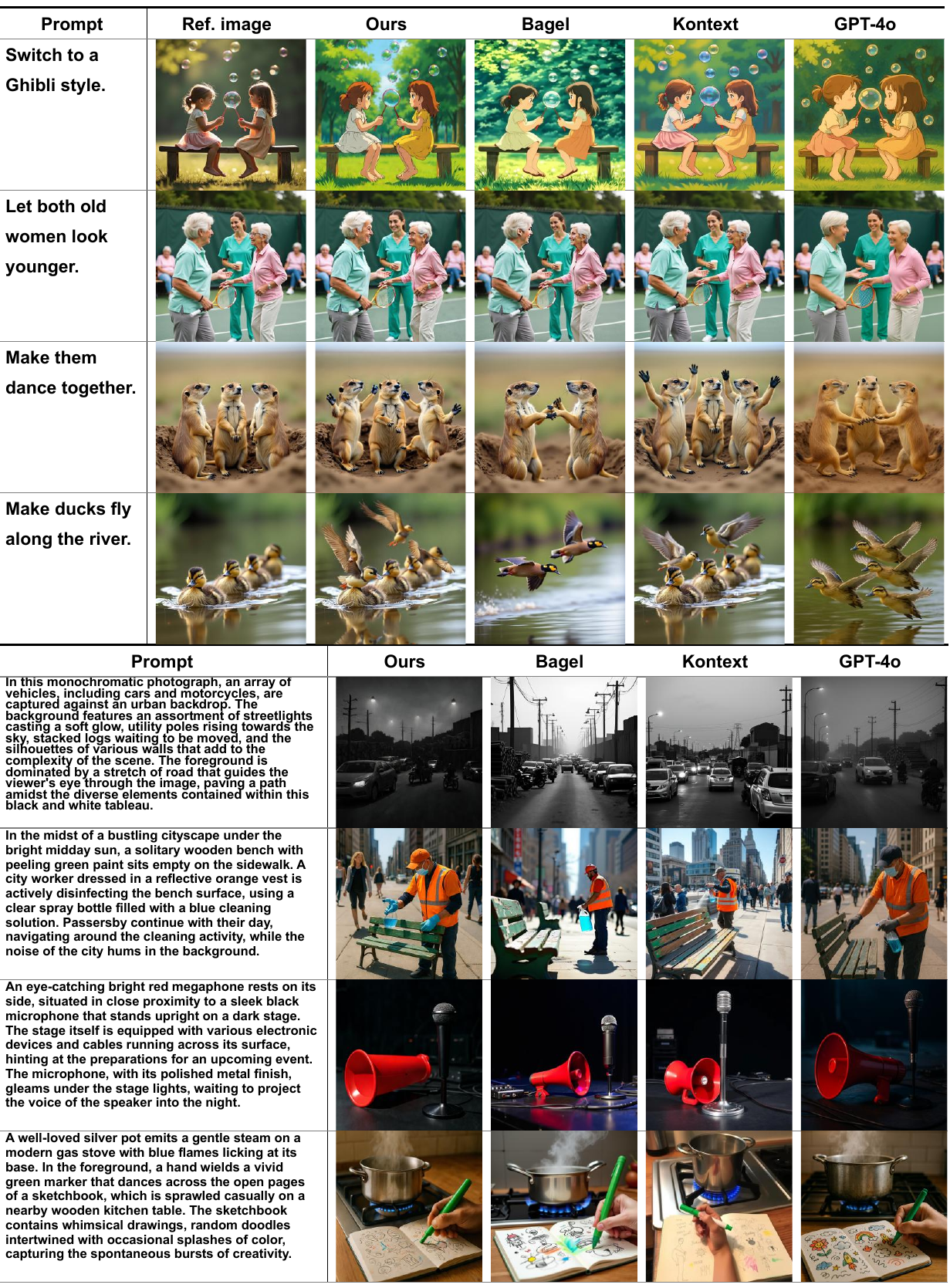}}
    \vspace{-1em}
    \caption{Failure cases.}
    \label{fig:failure_cases}
\end{figure}

\paragraph{Limitations.} While Skywork UniPic demonstrates strong performance across generation and editing tasks, certain limitations remain. As shown in Figure~\ref{fig:failure_cases}, the model occasionally struggles with complex or ambiguous instructions in text-to-image generation, leading to suboptimal instruction adherence. In the image editing setting, we observe cases where the model fails to respond to the editing prompt, resulting in incomplete or missing modifications. These limitations suggest that further refinement is needed in instruction grounding and editability robustness, particularly under challenging or compositional scenarios.

\paragraph{Emergence of Capabilities.} Similar to observations in BAGEL~\cite{deng2025emergingpropertiesunifiedmultimodal}, UniPic exhibits a clear, staged emergence of capabilities. Notably, text-to-image (T2I) generation appears in Stage 2 and is progressively refined, whereas more complex image editing capabilities emerge significantly later, only becoming evident in Stage 3 and Stage 4. This staggered manifestation reflects the inherent complexity of image editing, which demands a more sophisticated integration of visual-semantic alignment, conditional reasoning, and structural preservation compared to direct generation. In our work, we define an ability as emergent if it is absent in earlier training stages but materializes in later ones. This qualitative shift, often termed a phase transition, is consistent with our observation that UniPic's loss curves do not explicitly signal the onset of new capabilities, reinforcing the notion that training loss is an insufficient proxy for evaluating true model abilities.

To investigate this phenomenon, we evaluate model checkpoints from each stage by tracking average scores on standard VLM benchmarks (as a proxy for multimodal understanding), the GenEval score (for generation), and the GEdit-Bench performance (for editing). Our experiments consistently show that \textbf{editing capabilities emerge later than generation capabilities}, a pattern that holds even when scaling image resolution from $256 \times 256$ up to $1024 \times 1024$. Interestingly, each resolution increase induces a temporary performance dip followed by a rapid recovery that surpasses the previous capability plateau, suggesting higher resolutions unlock higher performance ceilings. Furthermore, we find no clear evidence that simply scaling understanding-centric data (e.g., image-text matching) directly enhances these generative or editing capabilities. This observation underscores the necessity of generation-specific training strategies for mastering complex, instruction-following tasks.

\clearpage
\section{Conclusion and Future Work}

We present Skywork UniPic, a unified autoregressive model that achieves competitive performance across image understanding, text-to-image generation, and image editing tasks within a single 1.5B parameter architecture. Through decoupled visual encoding that employs MAR for generation and SigLIP2 for understanding, our model resolves the fundamental tension between pixel-level fidelity and semantic understanding that has constrained previous unified approaches.

The model demonstrates strong empirical results: 0.86 on GenEval for compositional generation, 85.5 on DPG-Bench for complex instruction following, and 5.83 on GEdit-Bench for image editing, while maintaining efficient deployment on consumer hardware. Our comprehensive data construction pipelines address critical data scarcity in editing tasks, and the specialized reward modeling framework provides effective quality assurance for training data curation.

Key technical contributions include the decoupled encoding strategy that preserves both generation quality and understanding capabilities, systematic data construction methodologies for high-quality training corpus creation, and progressive training curriculum that enables efficient capability development across multiple resolutions. The work demonstrates that unified multimodal models can achieve both strong performance and practical efficiency, challenging assumptions about the necessity of massive parameter scaling for capable multimodal systems.

Future work will address current limitations including performance on highly complex compositional instructions, fine-grained editing precision in challenging scenarios, and further optimization for multilingual capabilities. The open-source release of model weights, training code, and datasets aims to facilitate further research in parameter-efficient unified multimodal architectures.

\section{Contributions}

\textbf{Core Contributors:} Peiyu Wang, Yi Peng, Yimeng Gan, Liang Hu, Eric Li*, Xuchen Song* \

\textbf{Contributors:} Tianyidan Xie, Xiaokun Wang, Yichen Wei, Chuanxin Tang, Bo Zhu, Changshi Li, Hongyang Wei, Yang Liu, Yahui Zhou

* Project Lead

\clearpage

\bibliography{main}
\bibliographystyle{plain}
\end{document}